%% file: v5.tex
\documentclass[letterpaper]{article} % DO NOT CHANGE THIS
\usepackage{aaai2027}
\usepackage[hyphens]{url}  % DO NOT CHANGE THIS
\usepackage{graphicx} % DO NOT CHANGE THIS
\usepackage{natbib}  % DO NOT CHANGE THIS AND DO NOT ADD ANY OPTIONS TO IT
\usepackage{caption} % DO NOT CHANGE THIS AND DO NOT ADD ANY OPTIONS TO IT
\usepackage{algorithm}
\usepackage{algorithmic}
\usepackage{amsmath}
\usepackage{amssymb}
\usepackage{booktabs}   % \toprule, \midrule, \bottomrule, \cmidrule
\usepackage{multirow}  % \multirow
\usepackage{pifont}    % \ding{55} for cross marks in tables
\usepackage{xcolor}    % colored check/cross marks
\usepackage{colortbl}  % row colors in tables
\newcommand{\cmark}{\textcolor{green!55!black}{\checkmark}}
\newcommand{\xmark}{\textcolor{red!80!black}{\ding{55}}}
\nocopyright

\title{MM-VeriAgent: Learning to Use Extensive Tools to Verify Multimodal Misinformation with Reinforcement Learning}

\author{Peipei Li$^{1}$   \quad
Shuhan Xia$^{1}$  \quad 
Shengyang Liu$^{1}$ \quad 
Zekun Li$^{2}$ \textsuperscript{\dag} \quad
Ran He$^{2,3}$ \quad
\\
$^{1}$ Beijing University of Posts and Telecommunications \quad \\
$^{2}$ Minzu University of China \quad \\
$^{3}$ NLPR, Institute of automation, Chinese Academy of Sciences \quad \\
\textsuperscript{\dag} Corresponding author.\quad \\
}

\affiliations{}

\newcommand{\toolset}{\mathcal{T}}

\newcommand{\sources}{\mathcal{S}}
\newcommand{\traj}{\tau}

\newcommand{\safeincludegraphics}[2]{%
  \IfFileExists{#2}{\includegraphics[#1]{#2}}{%
    \fbox{\parbox[c][5cm][c]{0.95\textwidth}{\centering \texttt{#2} (placeholder)}}}}

\begin{document}

\maketitle

\begin{abstract}
Real-world multimodal misinformation often involves mixed forgery sources, requiring sample-specific detection strategies. Existing tool-augmented methods rely on predefined workflows or inference-time planning, limiting adaptability or increasing inference cost.
To address this issue, we introduce \textbf{MM-VeriAgent}, which learns to verify mixed-source multimodal misinformation with tools. We first build \textbf{MM-VeriTools}, a specialized toolkit for misinformation detection agents. By benchmarking various candidate models and methods on the sub-tasks required by mixed-source detection, we select the strongest for textual, visual, and cross-modal forgery analysis and encapsulate them as callable tools with a unified interface. On top of this toolkit, we train the LVLM agent with reinforcement learning to teach it how to use these tools to better solve mixed-source detection. Since many of the tools are specialized models whose online execution at every rollout severely limits RL efficiency, we further introduce \textbf{Tool-Execution Cache}, which pre-executes candidate tool calls and reuses their cached outputs during training. This preserves multi-step rollouts while reducing online tool execution, largely improving the training efficiency.
Experiments on MMFakeBench demonstrate substantial accuracy gains over the base model without explicit tool search at inference time. Ablation and efficiency analyses further validate the learned tool-use policy and show that Tool-Execution Cache reduces online tool executions during training.
\end{abstract}

\begin{figure*}[t]
    \centering
    \safeincludegraphics{width=\textwidth}{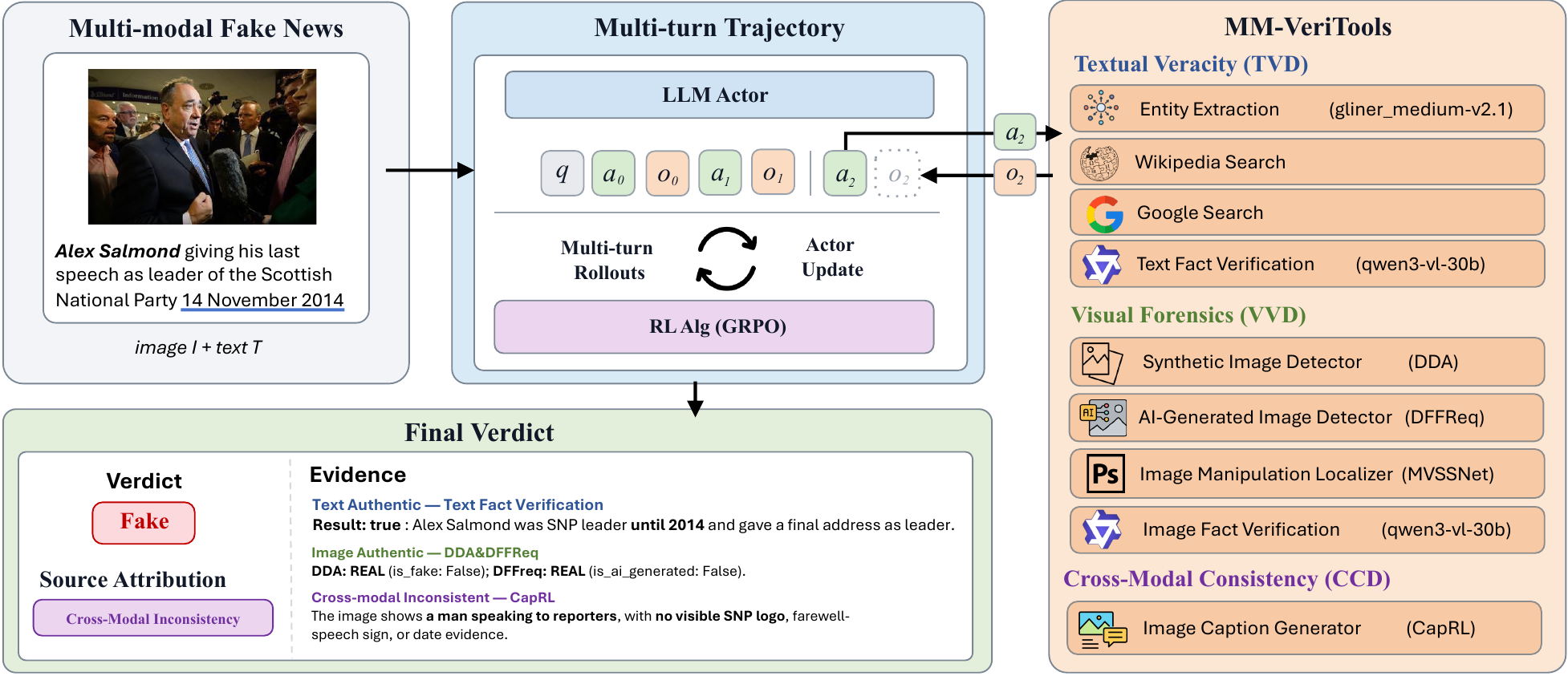}
    \caption{Overview of MM-VeriAgent. Given an image--text post $x=(I,T)$, the policy performs a multi-turn rollout: at each turn it either emits a tool-calling action $a_t$, which ends with a tool-specific stop token and invokes a tool from the MM-VeriTools toolkit (covering textual veracity, visual forensics, and cross-modal consistency), receiving the returned observation $o_t$, or terminates the interaction and outputs the final verdict together with fine-grained source attribution and supporting evidence. The policy is optimized with GRPO over multi-turn rollouts.}
    \label{fig:framework}
\end{figure*}

\section{Introduction}
Real-world multimodal misinformation rarely stems from a single type of forgery. As formalized by MMFakeBench~\cite{liu2024mmfakebench}, a deceptive image--text post may involve textual veracity distortion, visual veracity distortion, cross-modal consistency distortion, or combinations thereof, and the underlying source composition is unknown at test time. Crucially, different sources demand different verification mechanisms: fabricated claims call for fact checking against external knowledge, manipulated or AI-generated images call for visual forensics, and mismatched image--text pairs call for cross-modal consistency reasoning. A fixed detection pipeline is therefore inherently mismatched with mixed-source inputs. Effective detection requires deciding, for each sample, which evidence to collect and when to stop (Figure~\ref{fig:framework}).

Tool-augmented agents have recently emerged as a promising direction, coupling the reasoning ability of large vision-language models (LVLMs) with external retrieval and specialized detectors. Since MMD-Agent~\cite{liu2024mmfakebench} first cast mixed-source detection as a multi-turn tool-augmented verification process, this paradigm has been widely adopted~\cite{braun2024defame, beigi2024lrq, li2024large, cui2026t2agent}. Existing methods, however, orchestrate tools purely through prompting frozen LVLMs at inference time. Predefined workflows~\cite{liu2024mmfakebench, beigi2024lrq, li2024large} execute essentially the same tool sequence for every input and thus cannot adapt to sample-specific forgery sources. Test-time planning and search~\cite{braun2024defame, cui2026t2agent} adapt better, but every input incurs additional computation to plan subsequent calls or to evaluate candidate tool-use trajectories. Balancing sample-specific tool use with efficient inference therefore remains a key challenge.

To address this challenge, we first build \textbf{MM-VeriTools}, a specialized toolkit for misinformation detection agents. By benchmarking candidate models and methods on the sub-tasks required by mixed-source detection, we select the strongest for textual, visual, and cross-modal forgery analysis and encapsulate them as callable tools behind a unified interface. On top of this toolkit, we propose \textbf{MM-VeriAgent}. To our knowledge, it is the first model actually trained for mixed-source multimodal misinformation detection with reinforcement learning (RL), whereas prior agents perform direct inference. Optimized with task-specific rewards, the agent learns to decide which tool to invoke and when to stop, learning how to use these tools to better solve mixed-source detection.

However, optimizing the policy through online interaction with real tools is expensive. Many tools in the toolkit are specialized models trained for specific sub-tasks, so executing them is slow. Each RL rollout may invoke several such tools, and rollouts are repeated across many policy updates, making online tool execution the dominant bottleneck of policy optimization. We therefore introduce the \textbf{Tool-Execution Cache}. Before training, a high-capability teacher agent performs multiple rollouts for every training sample and records each executed tool call. During training, policy-generated calls are matched against the cache and answered with the recorded outputs. This shifts expensive tool execution from the RL loop to a one-off offline stage while preserving genuine multi-turn tool interaction.

% NOTE: number aligned with Table 1 (Qwen2.5-VL-7B multi-class ACC = 34.3); please verify against raw results.
On MMFakeBench, MM-VeriAgent raises the multi-class accuracy of its Qwen2.5-VL-7B backbone from 34.3\% to 70.4\%, on par with substantially larger proprietary models, without explicit tool search at inference time. Our main contributions are summarized as follows:
\begin{itemize}
\item We build MM-VeriTools, a specialized toolkit that benchmarks candidate models on the sub-tasks of mixed-source detection and encapsulates the strongest as callable tools behind a unified interface.
\item We propose MM-VeriAgent, to our knowledge the first model trained with RL for mixed-source multimodal misinformation detection, which learns to decide which tool to invoke and when to stop.
\item We introduce Tool-Execution Cache, which pre-executes tool calls and reuses their cached outputs during RL training, greatly reducing online tool execution.
\item Experiments on MMFakeBench show that MM-VeriAgent improves the multi-class accuracy of Qwen2.5-VL-7B from 34.3\% to 70.4\%, with ablation and efficiency analyses validating each component.
\end{itemize}

\section{Related work}
\label{sec:Related work}

\subsection{Misinformation Detection}
Early misinformation detection methods primarily focus on a single modality, either identifying factual distortions in textual claims~\cite{thorne2018fever, przybyla2020capturing, huang2023faking} or detecting manipulated and AI-generated visual content~\cite{wang2020cnn, ojha2023towards, liu2022detecting}. Real-world misinformation, however, is often conveyed through coupled image--text content, where deception may originate from textual distortion, visual manipulation, or inconsistency between the two modalities. MMFakeBench~\cite{liu2024mmfakebench} formalizes this mixed-source setting into three major forgery categories, providing a realistic benchmark for fine-grained misinformation verification.

Under this paradigm, some works train end-to-end multimodal classifiers that aggregate features across modalities~\cite{xu2025mdam3, papadopoulos2025red}, while others directly prompt or finetune LVLMs to identify misinformation from mixed sources~\cite{yan2025trust, cui2026t2agent, li2024large, lin2025fact, zeng2024multimodal}. Despite their effectiveness, most existing methods formulate multimodal verification as a single-shot prediction problem and thus cannot flexibly coordinate the heterogeneous expert capabilities required for textual fact checking, visual forensics, and cross-modal consistency analysis.

\input{tables/paradigm}
\subsection{Tool-Augmented Agents and Tool Learning}
Augmenting LLMs with external tools has become a general recipe for tasks that exceed the capability of a single forward pass. Early frameworks orchestrate tools without training. For example, OctoTools~\cite{lu2025octotools} encapsulates tools as standardized tool cards and coordinates them with a planner and an executor, but its tool selection relies on in-context reasoning alone and is static per task. Recent work instead trains tool use with reinforcement learning. Search-R1~\cite{jin2025search} learns when to issue search queries during reasoning, ReTool~\cite{feng2025retool} learns when and how to invoke a code interpreter, Tool-Star~\cite{dong2025toolstar} coordinates multiple tools with a hierarchical reward, and VerlTool~\cite{jiang2025verltool} provides a unified infrastructure for such agentic RL training. RL has also been introduced into fact verification, where an agent alternates between claim reasoning and evidence retrieval before predicting claim veracity~\cite{he2025veri}. However, these methods target textual reasoning tasks with homogeneous tool actions, and none addresses mixed-source multimodal verification, which requires routing each sample to heterogeneous forensic and retrieval tools. Table~\ref{tab:paradigm} contrasts MM-VeriAgent with existing tool-augmented detectors for this task: it is the only one that learns the tool-use policy and keeps inference free of planning and search.

A practical obstacle of such RL training is the cost of executing real tools inside rollouts. Prior remedies include caching API responses for stable evaluation~\cite{guo2024stabletoolbench}, simulating tool feedback with LLMs~\cite{li2025simia}, building an online prefix-tree cache during training~\cite{kumar2026tvcache}, and matching policy calls against cached rollouts with reward reweighting~\cite{islam2026cacherl}. Our Tool-Execution Cache instead constructs the cache offline via teacher rollouts and matches entries at the sample level, which suits read-only verification tools whose outputs are bound to individual image--text pairs.

\section{Method}

\subsection{Overview}

Figure~\ref{fig:framework} presents the overall framework of MM-VeriAgent. In the following, we first introduce the task and its multi-turn tool-use formulation (\textit{Task Definition}), and describe the construction of the \textbf{MM-VeriTools} toolkit (\textit{MM-VeriTools: Extensive Toolkit for Verification}). We then present how the agent is trained to use these tools (\textit{Reinforcement Learning for Verification}), and finally the \textbf{Tool-Execution Cache} that replaces online tool execution during training.

\subsection{Task Definition}

\paragraph{Mixed-source detection.} The input is an image--text post $x=(I,T)$, where the image $I$ and the text $T$ jointly present one piece of information and are expected to corroborate each other. The task is to predict a binary veracity label $y$ (real or fake) and a fine-grained source label $z \in \sources$, where $\sources$ consists of \textit{Real} (the post is authentic), \textit{Textual Veracity Distortion} (TVD, the text conveys false claims), \textit{Visual Veracity Distortion} (VVD, the image is manipulated or AI-generated), and \textit{Cross-modal Consistency Distortion} (CCD, the image and the text are semantically mismatched)~\cite{liu2024mmfakebench}. A post is fake once any distortion is present, and different distortions may co-occur in principle. Following MMFakeBench, each sample is annotated with a unique primary source label, so source attribution is a four-way classification.
% TODO: confirm how MMFakeBench labels samples with combined distortions (if any) among its 12 sub-categories.

\paragraph{Agentic Tool-Use for Verification.} Following tool-augmented agents~\cite{liu2024mmfakebench, cui2026t2agent}, we model verification as a multi-turn interaction between an agent $\pi_\theta$ and a toolset $\toolset$. At each turn, the agent generates an action $a_k$, a token sequence that either invokes a tool in $\toolset$ (terminated by a tool-specific stop token) or ends the interaction, yielding a trajectory
\begin{equation}
\traj =
\left(
a_0, o_0, a_1, o_1, \ldots, a_{K-1}, o_{K-1}, a_K
\right),
\label{eq:trajectory}
\end{equation}
where $o_k$ denotes the observation tokens returned by the invoked tool and appended to the context, and $K$ is the number of tool-calling turns. The final action $a_K$ outputs the verdict, which consists of the predicted veracity label $\hat{y}$, the predicted source label $\hat{z} \in \sources$, the supporting evidence $\hat{E}$, namely the set of evidence items cited from the collected tool observations to justify the decision, and a natural language reasoning.

\paragraph{Limitations of existing agents.} Existing agents realize this formulation by prompting frozen LVLMs over limited tools. The integrated tools cover only part of the required verification capabilities, and the prompted policy often invokes them redundantly or suboptimally. We address the two limitations with an extensive toolkit and a learned tool-use policy.

% NOTE: toolset construction figure removed (redundant with Table 2 and the text); can move to appendix if needed.
\iffalse
\begin{figure}[t]
    \centering
    \safeincludegraphics{width=\columnwidth}{fig/toolset.pdf}
    \caption{Construction pipeline of MM-VeriTools. \textbf{Step 1}: candidate specialized models and APIs are wrapped with a unified tool interface and organized into three capability groups (TVD, VVD, and CCD). \textbf{Step 2}: each candidate is benchmarked on the sub-task of its group, directly for off-the-shelf detection tools and through a single-tool agent for auxiliary tools, and the top performers are retained as the final toolset $\mathcal{T}$.}
    \label{fig:toolset}
\end{figure}
\fi

\subsection{MM-VeriTools: Extensive Toolkit for Verification}

Mixed-source detection requires heterogeneous verification capabilities, yet naively exposing every available tool to the policy introduces redundant evidence, a larger action space, and unnecessary execution overhead. We construct \textbf{MM-VeriTools} in two steps.

\input{tables/toolpool}
\paragraph{Step 1: Collection and grouping.} We collect candidate specialized models and APIs for the three verification sub-tasks, namely TVD, VVD, and CCD, and wrap them with a unified callable interface, as listed in Table~\ref{tab:toolpool}. Following MMD-Agent~\cite{liu2024mmfakebench} and T$^2$Agent~\cite{cui2026t2agent}, every tool takes fixed arguments from the queried sample, so a tool call specifies only the tool name. The TVD group examines the factual correctness of textual claims, the VVD group detects image manipulation and AI-generated content, and the CCD group assesses the semantic alignment between the image and the accompanying text.
% TODO: confirm the exact Qwen3-30B variant (VL? A3B?).

\paragraph{Step 2: Benchmarking and pruning.} Our principle is to admit only tools that help the base model or solve the sub-task better than the base model itself. To measure this, for each sub-task we build a balanced benchmark set from the training split of MMFakeBench, pairing 300 samples of the target source with 300 \textit{Real} samples and keeping it disjoint from the test set. The TVD, VVD, and CCD sets take the text, the image, and both as input, respectively. On each set we compare three types of candidates:
\begin{itemize}
\item \textit{Base}: GPT-5-mini~\cite{openai2025gpt5} without any tool, serving as the reference for selection.
\item \textit{Auxiliary}: tools that provide intermediate evidence, each attached to the base model as its only tool.
\item \textit{Direct}: tools whose outputs directly answer the sub-task, evaluated off the shelf.
\end{itemize}
A candidate is retained only if it outperforms the base model, a marginal-contribution criterion in the spirit of \cite{lu2025octotools}. The final toolset $\toolset$ contains nine tools, four textual, four visual, and one cross-modal, as marked in Table~\ref{tab:toolpool}.

Note that pruning only fixes the pool of available tools. Which tools to invoke for each sample is decided by the learned policy, in contrast to statically selecting tools per task~\cite{lu2025octotools}.

\subsection{Reinforcement Learning for Verification}

We train the agent with reinforcement learning to teach it which tool to invoke and when to stop. Under the formulation above, the trajectory distribution factorizes over the policy's own generations:
\begin{equation}
P_{\theta}(\traj \mid x)
=
\prod_{k=0}^{K}
\pi_{\theta}\!\left(a_k \mid x, a_{<k}, o_{<k}\right),
\label{eq:factorization}
\end{equation}
where the observations $o_k$ are injected by the environment rather than generated by the policy. We exploit this property both for cache-based training and for loss masking. MM-VeriAgent learns $\pi_\theta$ by maximizing the expected trajectory reward:
\begin{equation}
\max_{\theta}\;
\mathbb{E}_{x \sim \mathcal{D},\; \traj \sim P_{\theta}(\cdot \mid x)}
\left[ R(x, \traj) \right].
\label{eq:objective}
\end{equation}

\paragraph{Reward design.}
To guide the agent toward correct and evidence-grounded verification
procedures, we design a task-specific reward that evaluates the final
verdict along four aspects. The outcome rewards are indicator
functions, $R_{\mathrm{veracity}} = \mathbb{I}[\hat{y}=y]$ and
$R_{\mathrm{source}} = \mathbb{I}[\hat{z}=z]$, where $y$ and $z$ are
the ground-truth labels. The evidence reward
$R_{\mathrm{evi}} = |\hat{E}\cap E^{*}| \,/\, |\hat{E}\cup E^{*}|$
measures the overlap between the predicted evidence and the reference
evidence set $E^{*}$, addressing the case where the agent attributes
the correct source while citing irrelevant tool outputs. Such
fine-grained overlap rewards provide more informative learning signals
than binary indicators in tool learning~\cite{qian2025toolrl}. The
format reward
$R_{\mathrm{fmt}} = \mathbb{I}[\text{the verdict is parseable with
valid labels}]$ encourages well-formed outputs.
% TODO: define what \hat{E} and E^* are sets of (cache entry ids? evidence snippets?) per the actual implementation.
The final reward combines the four terms:
\begin{equation}
R
=\ \lambda_y R_{\mathrm{veracity}}
 + \lambda_z R_{\mathrm{source}} + \lambda_e R_{\mathrm{source}} R_{\mathrm{evi}}
 + \lambda_f R_{\mathrm{fmt}},
\label{eq:reward}
\end{equation}
where the evidence term is gated by $R_{\mathrm{source}}$, so that
evidence overlap is rewarded only under correct attribution. In
practice we assign larger weights to $\lambda_z$ and $\lambda_e$, as
fine-grained source attribution is the central objective of
mixed-source verification.

\paragraph{Cold-start SFT.}
The policy is first cold-started with
supervised finetuning (SFT) on verification demonstrations,
% TODO: confirm the composition of SFT demonstrations (e.g., teacher trajectories vs. annotated outputs),
% and whether tool observations are masked from the SFT loss (standard practice if demos contain tool outputs).
minimizing the negative log-likelihood
\begin{equation}
\mathcal{L}_{\mathrm{SFT}}(\theta)
=
-\,\mathbb{E}_{(x,\tau^{*})}
\left[\log \pi_{\theta}(\tau^{*} \mid x)\right],
\label{eq:sft}
\end{equation}
which equips the model with basic verification ability and the
tool-calling format.

\begin{figure*}[t]
    \centering
    \includegraphics[width=\textwidth]{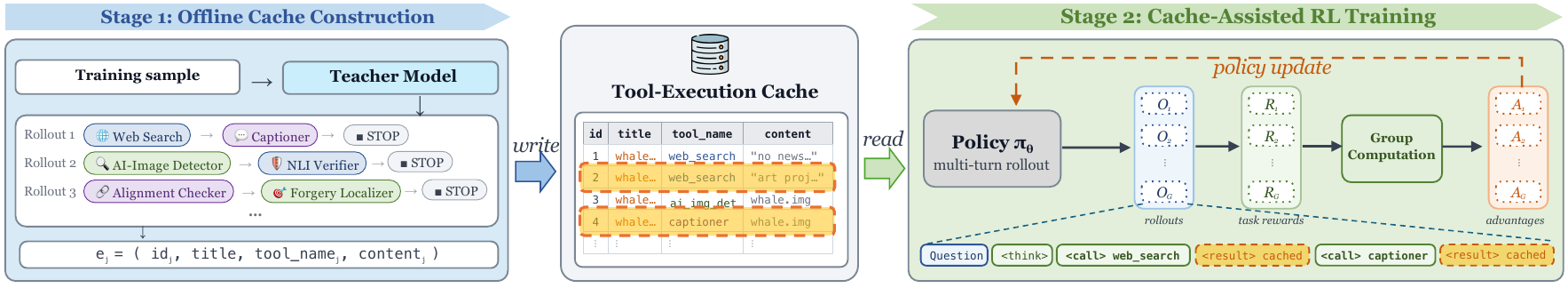}
    \caption{Overview of cache-assisted training. \textbf{Stage 1} constructs the Tool-Execution Cache offline via teacher rollouts, and \textbf{Stage 2} performs GRPO training with tool calls served from the cache.}
    \label{fig:cache}
\end{figure*}

\begin{figure}[t]
    \centering
    \includegraphics[width=\columnwidth]{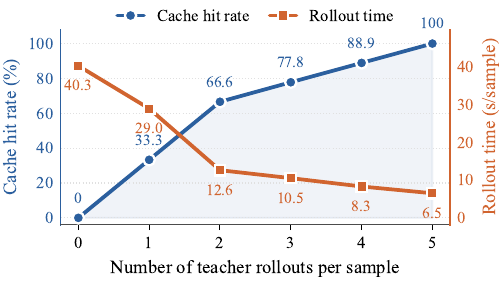}
    \caption{Cache hit rate and per-sample rollout time versus the number of teacher rollouts. Five rollouts yield a 100\% hit rate and cut the rollout time from 40.3 to 6.5 seconds.}
    \label{fig:cache_eff}
\end{figure}

\paragraph{GRPO training.}
The second stage optimizes the tool-use policy with
GRPO~\cite{shao2024deepseekmath}, which samples a group of $G$
trajectories $\{\tau_i\}_{i=1}^{G}$ from the old policy
$\pi_{\theta_{\mathrm{old}}}$ for each sample $x$ and computes
group-normalized advantages:
\begin{equation}
A_{i}
=
\frac{
R_{i}
-
\operatorname{mean}\!\left(\{R_{j}\}_{j=1}^{G}\right)
}{
\operatorname{std}\!\left(\{R_{j}\}_{j=1}^{G}\right)
}.
\label{eq:grpo_advantage}
\end{equation}

Different from standard GRPO, a trajectory here contains both action
tokens generated by the policy and observation tokens injected by
external tools. The observation tokens are off-policy with respect to
$\pi_{\theta}$ and are therefore excluded from the
loss~\cite{jin2025search}. Let
$\mathcal{A}_i$ denote the index set of action tokens in $\tau_i$. The
policy is updated with the token-level clipped objective
\begin{align}
\mathcal{L}_{\mathrm{clip}}(\theta)
=
\frac{1}{G}
\sum_{i=1}^{G}
\frac{1}{|\mathcal{A}_i|}
\sum_{t \in \mathcal{A}_i}
\min \Big(
&r_{i,t}(\theta) A_{i},
\nonumber\\
\operatorname{clip}\!\left(
r_{i,t}(\theta),
1-\epsilon,
1+\epsilon
\right) A_{i}
\Big),
\label{eq:grpo_clip}
\end{align}
where
$r_{i,t}(\theta)
=
\pi_{\theta}(\tau_{i,t}\mid\tau_{i,<t})
/
\pi_{\theta_{\mathrm{old}}}(\tau_{i,t}\mid\tau_{i,<t})$
is the token-level importance ratio. The overall training objective is
\begin{equation}
\mathcal{J}_{\mathrm{GRPO}}(\theta)
=
\mathbb{E}_{\substack{x \sim \mathcal{D} \\
\{\tau_i\} \sim \pi_{\theta_{\mathrm{old}}}(\cdot \mid x)}}
\big[
\mathcal{L}_{\mathrm{clip}}(\theta)
-
\beta D_{\mathrm{KL}}
(
\pi_{\theta}\,\|\,\pi_{\mathrm{ref}}
)
\big],
\label{eq:grpo_objective}
\end{equation}
where the KL term is likewise computed only over action tokens~\cite{jin2025search}.
In implementation, cached observations are tokenized separately and
concatenated with action segments, so the action tokens remain
identical to what the policy generated.

\subsection{Tool-Execution Cache}

Many tools in $\toolset$ are slow specialized models, and the RL stage repeats their execution across rollouts and policy updates, making online tool execution the dominant training bottleneck and a source of response variance. The Tool-Execution Cache decouples tool execution from policy optimization (Figure~\ref{fig:cache}).

\paragraph{Offline cache construction.} Before training, a teacher agent (GPT-5-mini at temperature 1.0) performs five rollouts per training sample with the pruned toolset, executing every tool call online and recording the returned context. Heterogeneous tool outputs, such as retrieved passages, forensic scores, and captions, are converted into a unified textual form and stored as entries
\begin{equation}
e_j =
\left(
\mathrm{id}_j,
\mathrm{title},
\mathrm{tool\_name}_j,
\mathrm{content}_j
\right),
\label{eq:cache_entry}
\end{equation}
where $\mathrm{title}$ identifies the image--text pair, $\mathrm{tool\_name}_j$ the invoked tool, $\mathrm{content}_j$ the recorded output, and $\mathrm{id}_j$ distinguishes entries from different rollouts. Multiple rollouts increase the coverage and diversity of the cached evidence.

\input{tables/main_results}
\input{tables/toolset_training}
\paragraph{Cache-based rollout.} During RL, when the policy invokes a tool, the environment matches the sample and the tool name against the cache and returns the cached content as the observation. Letting $\mathcal{C}_x$ denote the cache restricted to sample $x$, the observation at turn $k$ in Eq.~\eqref{eq:factorization} becomes
\begin{equation}
o_k = \mathcal{C}_x\!\left(\mathrm{tool}(a_k)\right),
\label{eq:cache_lookup}
\end{equation}
where $\mathrm{tool}(a_k)$ is the tool name parsed from $a_k$. The policy thus experiences full multi-turn tool interaction while cached calls execute no tool, and a cache miss falls back to online execution.

\paragraph{Validity of cached substitution.} Cached substitution is valid because every tool in $\toolset$ is stateless and takes fixed arguments: its output depends only on the invoked tool and the queried image--text pair. Sample-level matching therefore returns responses faithful to real execution, unlike stateful environments such as terminals, where a cached response is valid only if the entire call history matches~\cite{kumar2026tvcache}.

\paragraph{Efficiency.} As shown in Figure~\ref{fig:cache_eff}, the cache hit rate grows with the number of teacher rollouts per sample and reaches 100\% at five rollouts, where training requires no online tool execution and the rollout time drops from 40.3 to 6.5 seconds per sample, a 6.2$\times$ speedup.

% NOTE: Algorithm 1 commented out (redundant with Figure 3 + text); restore if reviewers ask for pseudocode.
\iffalse
\begin{algorithm}[t]
\caption{Cache-assisted training of MM-VeriAgent}
\label{alg:pipeline}
\begin{algorithmic}[1]
\REQUIRE training data $\mathcal{D}$, pruned toolset $\toolset$, teacher agent $\pi_{\mathrm{T}}$
\STATE \textit{// Stage 1: offline cache construction}
\FORALL{$x \in \mathcal{D}_{\mathrm{RL}}$}
    \STATE run $M$ rollouts of $\pi_{\mathrm{T}}$ with online tool execution
    \STATE record each executed call as an entry $e_j$ (Eq.~\ref{eq:cache_entry}) in cache $\mathcal{C}$
\ENDFOR
\STATE \textit{// Stage 2: cache-assisted policy optimization}
\STATE initialize $\pi_{\theta}$ via SFT (Eq.~\ref{eq:sft})
\WHILE{not converged}
    \STATE sample $G$ trajectories per instance, serving tool calls from $\mathcal{C}$ (Eq.~\ref{eq:cache_lookup})
    \STATE compute rewards (Eq.~\ref{eq:reward}) and advantages (Eq.~\ref{eq:grpo_advantage})
    \STATE update $\theta$ with the masked GRPO objective (Eq.~\ref{eq:grpo_objective})
\ENDWHILE
\end{algorithmic}
\end{algorithm}
\fi

\section{Experiments}

\subsection{Experimental Setup}

\paragraph{Datasets and metrics.}
We evaluate MM-VeriAgent on the mixed-source multimodal misinformation benchmark MMFakeBench. The dataset contains 11,000 image-text pairs and categorizes news into four classes: \textit{Real}, \textit{Textual Veracity Distortion (TVD)}, \textit{Visual Veracity Distortion (VVD)}, and \textit{Cross-modal Consistency Distortion (CCD)}.
We use 10,000 samples for training and 1,000 samples for testing.
Accordingly, we evaluate from two perspectives: binary veracity prediction (real vs.\ fake), for which we report accuracy and F1, and fine-grained source attribution, a multi-class problem over the four classes for which we report accuracy and macro-F1. The latter weighs all classes equally and thus remains informative under class imbalance.

% NOTE: Gemini-3-Flash/Pro-Preview rows are commented out in tables/main_results.tex;
% re-add them here if those rows are restored.
\paragraph{Baselines.}
We compare MM-VeriAgent with two families of baselines.
The first family, denoted \textit{Direct Prompting}, predicts with a single forward pass without any tool or agentic loop, covering both proprietary models~\cite{comanici2025gemini25, openai2024gpt4o, openai2025gpt5, openai2026gpt55} and open-source models~\cite{bai2025qwen25vl, bai2025qwen3vl}.
The second family consists of methods specialized for multimodal misinformation detection, including the agentic MMD-Agent~\cite{liu2024mmfakebench} and T$^2$Agent~\cite{cui2026t2agent} as well as the preference-optimized SCPO~\cite{gao2026thinking}. More details are provided in the supplementary material.

\paragraph{Implementation details.}
MM-VeriAgent is built upon Qwen2.5-VL-7B~\cite{bai2025qwen25vl} and trained with a two-stage pipeline consisting of supervised fine-tuning (SFT) followed by reinforcement learning (RL). The SFT stage is performed on 8,000 training samples, while the remaining 2,000 samples are used for RL policy optimization. All experiments are conducted on 8 NVIDIA H20 GPUs with a batch size of 16. The maximum number of tool-calling turns is set to 3. During RL training, we adopt Group Relative Policy Optimization (GRPO) and sample $n=4$ responses for each query. The reward weights are set to $\lambda_y=\lambda_f=0.3$ and $\lambda_z=\lambda_e=2.0$, assigning larger rewards to fine-grained source attribution and evidence-grounded verification.

\subsection{Main Results}

Table~\ref{tab:main_results} compares MM-VeriAgent with both families of baselines. We draw two key observations.

\textbf{Mixed-source verification remains challenging for existing MLLMs.} Even the strongest proprietary models stay around 70\% on multi-class source attribution (69.8\% for Gemini-2.5-Pro and 71.8\% for GPT-5.5), confirming that the task demands more than a single forward prediction.

\textbf{MM-VeriAgent delivers proprietary-level verification with a 7B backbone.} Trained on the same Qwen2.5-VL-7B, MM-VeriAgent raises the multi-class accuracy from 34.3\% to 70.4\% and achieves the best binary F1 (85.1\%) overall, outperforming MMD-Agent and SCPO on both tasks. T$^2$Agent reports a higher multi-class accuracy, but it relies on the proprietary GPT-4o with Monte Carlo tree search at test time, whereas MM-VeriAgent executes a single learned policy without any test-time search.

\begin{figure}[t]
    \centering
    \includegraphics[width=\columnwidth]{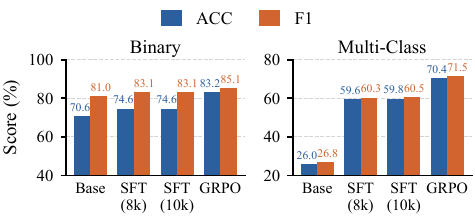}
    \caption{Effect of training with the toolset fixed to MM-VeriTools. Adding the same 2k samples as supervised data (SFT-10k) barely helps, whereas GRPO unlocks the toolkit.}
    \label{fig:training_ladder}
\end{figure}
\subsection{Effect of Toolset and Training}

\paragraph{Effect of the toolset.} Table~\ref{tab:toolset_training} varies the method on two untrained models. First, tools alone do not help a small model. The untrained Qwen2.5-VL-7B gains moderately with the compact MMD-Agent toolset (34.3\% to 41.3\% on multi-class accuracy), but degrades below direct prompting with our toolset (26.0\%), as the richer action space overwhelms an untrained policy.
Second, prompting cannot fully exploit an extensive toolkit even with a strong model. GPT-5-mini benefits from tools in general, yet performs better with the MMD-Agent toolset (65.2\%) than with ours (57.1\%).

\paragraph{Effect of training.} Figure~\ref{fig:training_ladder} fixes the toolset to MM-VeriTools and varies the training of Qwen2.5-VL-7B. SFT on the first 8k rollouts already lifts the multi-class accuracy from 26.0\% to 59.6\%. The remaining 2k samples then isolate the effect of reinforcement learning from that of additional data: adding them as more supervised data (SFT-10k) barely moves the accuracy to 59.8\%, whereas optimizing on them with GRPO reaches 70.4\%. The gain therefore comes from interaction-based optimization rather than from seeing more data.
Moreover, the trained agent surpasses its teacher. GPT-5-mini generates the rollouts for our cold-start SFT data and the Tool-Execution Cache, but reaches only 57.1\% with the same toolkit (Table~\ref{tab:toolset_training}), indicating that the agent learns a genuine tool-use policy rather than merely distilling the teacher.

\input{tables/ablation}
\subsection{Ablation Study}

\paragraph{Verification tools.} Table~\ref{tab:ablation} (top) ablates the three tool categories. Removing any category degrades both binary and multi-class performance, and removing CCD tools hurts most, dropping the multi-class accuracy from 70.4\% to 58.3\% and the macro-F1 from 71.5\% to 58.6\%. Cross-modal distortion cannot be identified from either modality alone, so without explicit cross-modal evidence the agent frequently confuses different misinformation sources. The three categories thus provide complementary evidence that the learned policy composes per sample.

\paragraph{Evidence-grounded reward.} Table~\ref{tab:ablation} (bottom) removes the evidence-grounded reward. Performance degrades on both tasks, most visibly on multi-class classification, where the accuracy and macro-F1 drop from 70.4\% to 66.1\% and from 71.5\% to 66.7\%. Optimizing prediction correctness alone is thus insufficient. Rewarding evidence overlap steers the agent toward source-consistent evidence, on which fine-grained attribution relies.

\section{Conclusion}

We presented MM-VeriAgent, which learns to verify mixed-source multimodal misinformation with extensive tools. Mixed-source verification requires a model to learn not only \emph{what} to predict but \emph{how} to verify. MM-VeriAgent instantiates this view with a benchmarked toolkit, reinforcement learning that teaches the agent which tool to invoke and when to stop, and a Tool-Execution Cache that keeps the training cost practical. On MMFakeBench, the learned policy more than doubles the multi-class accuracy of its 7B backbone without any test-time search. We hope this work moves multimodal misinformation detection from answer-only label prediction toward auditable, evidence-grounded verification.

\section{Limitations}

MM-VeriAgent inherits the quality of its tools: if retrieval fails or a forensic detector is miscalibrated, the agent may accumulate misleading evidence. The Tool-Execution Cache relies on stateless tools with sample-determined inputs, and tools with free-form arguments or time-varying outputs would require finer-grained matching and cache versioning. Our evaluation is also confined to MMFakeBench, currently the primary benchmark for mixed-source verification, so generalization to unseen forgery generators, news domains, and distortion types beyond its three source categories remains to be established. Extending the toolkit to a new distortion type further requires re-benchmarking candidate tools and re-training the policy. Finally, tool-based verification is more expensive than answer-only classification, so the tool budget should scale with risk.

\bibliography{aaai2027}

\end{document}

%% file: tables/paradigm.tex
% Comparison of tool-augmented MMD detectors
\begin{table}[t]
    \centering
    \caption{Comparison of tool-augmented detectors for mixed-source
    multimodal misinformation detection. Cache denotes an offline
    tool-execution cache for training.}
    \label{tab:paradigm}
    \setlength{\tabcolsep}{4pt}
    \renewcommand{\arraystretch}{1.12}
    \resizebox{\columnwidth}{!}{
    \begin{tabular}{llcccc}
        \toprule
        \textbf{Method} & \textbf{Organization}
        & \textbf{Adaptive} & \textbf{Search-free}
        & \textbf{Learned} & \textbf{Cache} \\
        \midrule
        MMD-Agent    & Fixed workflow     & \xmark & \cmark & \xmark & \xmark \\
        LRQ-FACT     & Fixed workflow     & \xmark & \cmark & \xmark & \xmark \\
        DEFAME       & Test-time planning & \cmark & \xmark & \xmark & \xmark \\
        T$^2$Agent   & Test-time search   & \cmark & \xmark & \xmark & \xmark \\
        \midrule
        \textbf{Ours} & \textbf{RL-learned policy}
        & \cmark & \cmark & \cmark & \cmark \\
        \bottomrule
    \end{tabular}
    }
\end{table}

%% file: tables/toolpool.tex
% Candidate tool pool with benchmarking results
\begin{table*}[t]
    \centering
    \caption{Candidate tool pool and group-wise benchmarking of
    MM-VeriTools. Each auxiliary tool is attached to the base model,
    and direct tools are evaluated off the shelf. A candidate is
    selected only if it outperforms the base model (Acc in
    \textbf{bold}).}
    \label{tab:toolpool}
    \setlength{\tabcolsep}{6pt}
    \renewcommand{\arraystretch}{1.12}
    \small
    \begin{tabular}{lllccc}
        \toprule
        \textbf{Group} & \textbf{Model / Method} & \textbf{Role}
        & \textbf{Type} & \textbf{Acc} & \textbf{Selected} \\
        \midrule
        \multirow{6}{*}{TVD}
        & \cellcolor{gray!12}GPT-5-mini & \cellcolor{gray!12}Base model & \cellcolor{gray!12}Base & \cellcolor{gray!12}0.63 & \cellcolor{gray!12}-- \\
        & GLiNER-medium~\cite{zaratiana2024gliner} & Entity extraction & Auxiliary & \textbf{0.71} & \cmark \\
        & Wikipedia search & Knowledge retrieval & Auxiliary & \textbf{0.71} & \cmark \\
        & Google search & Web retrieval & Auxiliary & \textbf{0.75} & \cmark \\
        & Qwen3-30B~\cite{yang2025qwen3} & Text fact verification & Direct & \textbf{0.69} & \cmark \\
        & Faking Fake News~\cite{huang2023faking} & Fake news detection & Direct & 0.35 & \xmark \\
        \midrule
        \multirow{6}{*}{VVD}
        & \cellcolor{gray!12}GPT-5-mini & \cellcolor{gray!12}Base model & \cellcolor{gray!12}Base & \cellcolor{gray!12}0.66 & \cellcolor{gray!12}-- \\
        & DFFreq~\cite{yan2026dffreq} & AI-generated image detection & Direct & \textbf{0.94} & \cmark \\
        & DDA~\cite{chen2025dda} & Synthesis and editing detection & Direct & \textbf{0.91} & \cmark \\
        & MVSS-Net~\cite{chen2021mvss} & Manipulation localization & Direct & \textbf{0.86} & \cmark \\
        & Qwen3-30B~\cite{yang2025qwen3} & Manipulation description & Direct & \textbf{0.80} & \cmark \\
        & UnivFD~\cite{ojha2023towards} & AI-generated image detection & Direct & 0.31 & \xmark \\
        \midrule
        \multirow{4}{*}{CCD}
        & \cellcolor{gray!12}GPT-5-mini & \cellcolor{gray!12}Base model & \cellcolor{gray!12}Base & \cellcolor{gray!12}0.53 & \cellcolor{gray!12}-- \\
        & CapRL~\cite{xing2026caprl} & Image captioning & Auxiliary & \textbf{0.92} & \cmark \\
        & HAMMER~\cite{shao2023dgm4} & Manipulation grounding & Direct & 0.49 & \xmark \\
        & FKA-Owl~\cite{liu2024fka} & Consistency detection & Direct & 0.44 & \xmark \\
        \bottomrule
    \end{tabular}
\end{table*}

%% file: tables/main_results.tex
% Required packages:
% \usepackage{booktabs}
% \usepackage{multirow}

\begin{table*}[t]
    \centering
    \caption{
        Main results on MMFakeBench.
        Proprietary and open-source MLLMs predict with a single
        forward pass without any tool, and specialized methods
        are executed with their original frameworks.
        All results are reported in percentage (\%) and the best
        result in each metric is in \textbf{bold}.
    }
    \label{tab:main_results}

    \setlength{\tabcolsep}{10pt}
    \renewcommand{\arraystretch}{1.15}

    \begin{tabular}{llcccc}
        \toprule
        \multirow{2}{*}{\textbf{Method}}
        & \multirow{2}{*}{\textbf{Model}}
        & \multicolumn{2}{c}{\textbf{Binary Classification}}
        & \multicolumn{2}{c}{\textbf{Multi-Class Classification}} \\
        \cmidrule(lr){3-4}
        \cmidrule(lr){5-6}
        & &
        \textbf{ACC}
        & \textbf{F1}
        & \textbf{ACC}
        & \textbf{F1} \\
        \midrule

        \multicolumn{6}{l}{\textit{Proprietary MLLMs}} \\
        Direct Prompting & Gemini-2.5-Flash & 78.8 & 78.4 & 60.5 & 60.2 \\
        Direct Prompting & Gemini-2.5-Pro & \textbf{86.7} & 83.2 & 69.8 & 69.5 \\
        % -- & Gemini-3-Flash-Preview & 90.4 & 88.4 & 70.9 & 69.5 \\
        % -- & Gemini-3-Pro-Preview & 90.1 & 88.2 & 74.2 & 72.6 \\
        Direct Prompting & GPT-4o & 74.0 & 76.8 & 60.9 & 49.2 \\
        Direct Prompting & GPT-5-mini & 78.6 & 76.1 & 53.4 & 50.6 \\
        Direct Prompting & GPT-5.5 & 80.6 & 79.5 & 71.8 & 69.0 \\

        \midrule

        \multicolumn{6}{l}{\textit{Open-source MLLMs}} \\
        Direct Prompting & Qwen2.5-VL-3B-Instruct & 48.7 & 48.6 & 38.9 & 28.1 \\
        Direct Prompting & Qwen2.5-VL-7B-Instruct & 72.8 & 53.9 & 34.3 & 25.0 \\
        Direct Prompting & Qwen3-VL-4B-Instruct & 75.4 & 70.4 & 43.5 & 29.6 \\
        Direct Prompting & Qwen3-VL-8B-Instruct & 77.6 & 68.6 & 49.1 & 39.3 \\

        \midrule

        \multicolumn{6}{l}{\textit{Specialized methods}} \\
        MMD-Agent & GPT-4V & 76.8 & 74.0 & 62.1 & 61.6 \\
        T$^{2}$Agent & GPT-4o & -- & -- & \textbf{75.3} & \textbf{75.9} \\
        SCPO & Qwen2-VL-7B & -- & -- & 66.0 & 66.1 \\

        \midrule

        \textbf{MM-VeriAgent (Ours)} & Qwen2.5-VL-7B & 83.2 & \textbf{85.1} & 70.4 & 71.5 \\

        \bottomrule
    \end{tabular}

\end{table*}

%% file: tables/toolset_training.tex
% Effect of toolset on untrained models
\begin{table}[t]
    \centering
    \caption{Effect of the toolset on untrained models. Prompted
    policies exploit the compact MMD-Agent toolset better than the
    extensive MM-VeriTools, which can even hurt the small model.}
    \label{tab:toolset_training}
    \setlength{\tabcolsep}{4pt}
    \renewcommand{\arraystretch}{1.05}
    \small
    \resizebox{\columnwidth}{!}{
    \begin{tabular}{llcccc}
        \toprule
        \multirow{2}{*}{\textbf{Model}}
        & \multirow{2}{*}{\textbf{Method}}
        & \multicolumn{2}{c}{\textbf{Binary}}
        & \multicolumn{2}{c}{\textbf{Multi-Class}} \\
        \cmidrule(lr){3-4}
        \cmidrule(lr){5-6}
        & & \textbf{ACC} & \textbf{F1} & \textbf{ACC} & \textbf{F1} \\
        \midrule
        \multirow{3}{*}{Qwen2.5-VL-7B}
        & Direct Prompting & 72.8 & 53.9 & 34.3 & 25.0 \\
        & MMD-Agent & 72.5 & 71.7 & 41.3 & 43.5 \\
        & Ours & 70.6 & 81.0 & 26.0 & 26.8 \\
        \midrule
        \multirow{3}{*}{GPT-5-mini}
        & Direct Prompting & 78.6 & 76.1 & 53.4 & 50.6 \\
        & MMD-Agent & 82.1 & 79.6 & 65.2 & 65.5 \\
        & Ours & 85.7 & 83.2 & 57.1 & 55.2 \\
        \bottomrule
    \end{tabular}
    }
\end{table}

%% file: tables/ablation.tex
% Merged ablation table: tool groups (top) + evidence reward (bottom)
\begin{table}[t]
    \centering
    \caption{Ablation studies. Top: removing each category of
    verification tools (TVD: textual verification; VVD: visual
    forensics; CCD: cross-modal consistency). Bottom: removing the
    evidence-grounded reward. All results in \%.}
    \label{tab:ablation}
    \setlength{\tabcolsep}{8pt}
    \renewcommand{\arraystretch}{1.12}

    \begin{tabular}{lcccc}
        \toprule
        \multirow{2}{*}{\textbf{Variant}}
        & \multicolumn{2}{c}{\textbf{Binary}}
        & \multicolumn{2}{c}{\textbf{Multi-Class}} \\
        \cmidrule(lr){2-3}
        \cmidrule(lr){4-5}
        & \textbf{ACC} & \textbf{F1}
        & \textbf{ACC} & \textbf{F1} \\
        \midrule
        w/o TVD tools
        & 81.7 & 82.1 & 63.1 & 64.3 \\
        w/o VVD tools
        & 82.2 & 82.6 & 67.2 & 67.9 \\
        w/o CCD tools
        & 79.9 & 81.4 & 58.3 & 58.6 \\
        \midrule
        w/o evidence reward
        & 77.1 & 84.8 & 66.1 & 66.7 \\
        \midrule
        \textbf{MM-VeriAgent}
        & \textbf{83.2} & \textbf{85.1}
        & \textbf{70.4} & \textbf{71.5} \\
        \bottomrule
    \end{tabular}
\end{table}